\documentclass[letterpaper, 10 pt, conference]{ieeeconf}  

\IEEEoverridecommandlockouts                              

\usepackage{graphics} 
\DeclareGraphicsExtensions{.pdf,.png,.jpg}
\usepackage{epsfig} 
\usepackage{times} 
\usepackage{amsmath} 
\usepackage{amssymb}  
\usepackage{kotex}

\usepackage{caption}
\usepackage{subcaption}
\usepackage{changepage}
\usepackage{xcolor}
\usepackage{multirow}
\usepackage{booktabs}

\usepackage{pifont}
\usepackage{colortbl}
\usepackage{graphicx} 

\newcommand{\cmark}{\textcolor{green!70!black}{\ding{51}}}
\newcommand{\xmark}{\textcolor{red}{\ding{55}}}

\definecolor{best}{HTML}{B4E1B9}   
\definecolor{second}{HTML}{FFF2AA} 
\definecolor{third}{HTML}{FCD7BA}  
\definecolor{graytext}{HTML}{808080} 

\newcommand{\best}[1]{\cellcolor{best}\textbf{#1}}
\newcommand{\second}[1]{\cellcolor{second}#1}
\newcommand{\third}[1]{\cellcolor{third}#1}

\title{\LARGE \bf
BayesianGS-SLAM: Uncertainty-Aware Neural Rendering SLAM via Probabilistic Formulation
}

\author{Kyeongsu Kang, Seongbo Ha, Sibaek Lee and Hyeonwoo Yu$^{*}$ 
 \thanks{*Corresponding author.}
 \thanks{Kyeongsu Kang, Seongbo Ha, Sibaek Lee and Hyeonwoo Yu are with the Department of Intelligent Robotics, Sungkyunkwan University, Suwon, South Korea. {\tt\small \{thithin0821, sobo3607, lmjlss, hwyu\}@skku.edu}. }
}

\begin{document}

\maketitle
\thispagestyle{empty}
\pagestyle{empty}


\begin{abstract}


Neural-rendering-based SLAM relies on rendered RGB-D residuals for camera tracking and map optimization, but the reliability of these predictions can vary substantially because of sensor noise, limited observation coverage, and incomplete map representations. Without an explicit reliability estimate, unreliable residuals may adversely affect pose optimization, while frames already well explained by the current map may trigger redundant mapping updates. In this paper, we present BayesianGS-SLAM, an uncertainty-aware 3D Gaussian Splatting SLAM framework that estimates predictive color and depth uncertainty during mapping and consistently reuses it across the SLAM pipeline. Our tractable probabilistic formulation combines a sensor-noise uncertainty component with an opacity-induced map-representation component propagated through the rendering process. The resulting predictive uncertainty is used to augment mapping, normalize tracking residuals through a robust pose objective, and evaluate incoming frames using a predictive-surprise-based keyframe criterion. Unlike prior uncertainty-aware neural-rendering SLAM methods that primarily consider color uncertainty or use uncertainty only during mapping, our framework estimates predictive uncertainty for both color and depth and integrates it into mapping, tracking, and keyframe selection. Evaluations on real-world RGB-D datasets demonstrate substantially improved depth uncertainty--error ranking compared with existing uncertainty-aware SLAM methods. Moreover, the proposed keyframe-selection strategy reduces the number of selected keyframes and mapping calls while maintaining competitive tracking and rendering performance.

\end{abstract}

\section{Introduction}


With the emergence of 3D reconstruction methods based on neural rendering, substantial progress has been made not only in computer vision but also in robotics, with particularly notable advances in SLAM. 
Neural-rendering-based SLAM has introduced diverse scene representations, ranging from continuous implicit neural fields to explicit differentiable primitives such as 3D Gaussians, enabling direct image-based optimization and high-quality rendering \cite{sucar2021imap, zhu2022nice}.
Furthermore, the advent of 3D Gaussian Splatting has accelerated the development of neural rendering-based SLAM, including 3DGS-based SLAM frameworks and systems with loop closure capabilities~\cite{keetha2024splatam,matsuki2024gaussian,zhu2025loopsplat,liso2024loopy, ha2024rgbd}.



Despite these advances, neural rendering-based SLAM still has several limitations. One of the most fundamental issues is confidence estimation, or uncertainty quantification. Although classical SLAM is fundamentally rooted in Bayesian estimation, neural rendering-based SLAM methods have received relatively limited attention from the perspectives of statistical inference and uncertainty modeling. Recent studies have explored uncertainty quantification for active mapping or active learning of keyframes \cite{yan2023active, pan2022activenerf}; however, these efforts are mostly confined to the mapping stage or only locally integrated into SLAM, and the development of a unified neural rendering-based SLAM framework remains limited. 


This limitation can be attributed to two main factors. The first is the substantial computational cost required for uncertainty quantification in neural rendering-based SLAM \cite{gal2016dropout}. In neural models, uncertainty quantification is challenging because their nonlinearity makes distributional inference difficult \cite{kendall2017uncertainties}. To address this issue, various uncertainty quantification methods have been introduced; however, most of them either train multiple networks simultaneously to measure the variance of their outputs \cite{lakshminarayanan2017simple}, infer uncertainty through sampling-based approaches \cite{shen2022conditional, shen2021stochastic}, or perform inference on already trained parameters or neural fields \cite{goli2024bayes}. Consequently, these methods may increase the computational cost or fail to be tightly coupled with SLAM.


Another practical challenge is that uncertainty estimates must remain useful for downstream SLAM optimization. Since uncertainty directly affects residual weighting and frame selection, its representation and integration into the SLAM pipeline are important for maintaining robust online operation.

In this paper, we propose an uncertainty-aware neural-rendering SLAM framework that estimates predictive RGB-D uncertainty during mapping and consistently reuses it across mapping, tracking, and keyframe selection. The predictive uncertainty combines a sensor-noise uncertainty component with an opacity-induced map-representation uncertainty component propagated through rendering linearization. The resulting uncertainty augments the mapping objective and is reused for uncertainty-normalized robust tracking. For keyframe selection, we use the negative posterior-predictive likelihood as a predictive-surprise score motivated by information gain. The proposed framework reduces redundant mapping updates while maintaining competitive tracking and rendering performance.

Our contributions are summarized as follows:
\begin{itemize}
    \item 
    We introduce a tractable predictive-uncertainty formulation for RGB-D neural rendering that combines sensor-noise and map-representation uncertainty for both color and depth.
    \item 
    We integrate the same predictive uncertainty into uncertainty-aware mapping, uncertainty-normalized tracking, and predictive-surprise-based keyframe selection, reducing redundant mapping updates while maintaining competitive SLAM performance.
\end{itemize}

\section{Related work}

\subsection{Uncertainty Quantification in neural rendering}

Research on uncertainty in neural rendering, including NeRF and 3DGS, has been actively pursued since the emergence of neural rendering techniques. These studies span a broad range, from methods that consider uncertainty only in image rendering to approaches that separately model aleatoric and epistemic uncertainty  \cite{gal2016dropout, kendall2017uncertainties}. For example, \cite{martin2021nerf} leveraged uncertainty to achieve more accurate rendering in outdoor scenes, while \cite{pan2022activenerf} considered color uncertainty and implemented active learning for frame selection by rendering uncertainty maps. However, as discussed in~\cite{shen2022conditional}, these methods differ in how uncertainty is represented and propagated through the rendering process, and some formulations provide only an indirect connection between the modeled uncertainty and the final rendered prediction.
In contrast, studies that separately model epistemic and aleatoric uncertainty in 3D space within neural rendering have typically borrowed techniques from Bayesian neural networks to quantify uncertainty \cite{shen2022conditional}.


Other Recent studies have developed uncertainty- and information-related formulations that are applicable to 3DGS. FisherRF~\cite{jiang2023fisherrf} uses Fisher-information-derived scores for uncertainty quantification and informative-view selection. PUP 3D-GS~\cite{hanson2025pup} derives a second-order spatial sensitivity score, using a Fisher approximation of the reconstruction-error Hessian, to identify Gaussians that can be pruned with limited impact on reconstruction quality. POp-GS~\cite{wilson2025pop} approximates the covariance of 3DGS parameters and formulates information-based view selection through optimal experimental design. These Magic Moments~\cite{ewen2025these} instead computes higher-order moments directly from the probabilistic rendering process, providing uncertainty estimates for color, depth, and semantic outputs without additional uncertainty-specific training. Recent methods have also modeled geometric or visibility-related uncertainty, including the spatial uncertainty field of UNG-GS~\cite{tan2025uncertainty} and the anisotropic visibility field of GAVIS~\cite{xue2026uncertainty}. Collectively, these approaches illustrate that uncertainty-related quantities in neural rendering may capture different properties, including parameter sensitivity, predictive reliability, geometric ambiguity, observation coverage, and active learning \cite{jiang2024ag}.

\subsection{Uncertainty-based neural rendering SLAM}


Recent neural-rendering SLAM methods have incorporated uncertainty into tracking and mapping. UNCLE-SLAM \cite{sandstrom2023uncle} learns observation-dependent depth uncertainty for NeRF-based SLAM, while CG-SLAM \cite{hu2024cg} incorporates depth uncertainty into 3DGS mapping. VBGS-SLAM \cite{zhu2026vbgs} introduces probabilistic Gaussian parameters through variational inference, and VarSplat \cite{tran2026varsplat} learns rendered color uncertainty for uncertainty-weighted tracking and mapping. Related uncertainty and information measures have also been used in active SLAM to select future views or trajectories for improving localization and mapping. 

In contrast, our method combines sensor-noise uncertainty with map-representation uncertainty for both color and depth, and reuses the resulting predictive uncertainty across mapping, uncertainty-normalized tracking, and predictive-surprise-based keyframe selection from an already observed RGB-D stream

\section{Method}
\begin{figure*}[tbh!]
    \centering
    \includegraphics
    [width=0.99\textwidth]
    {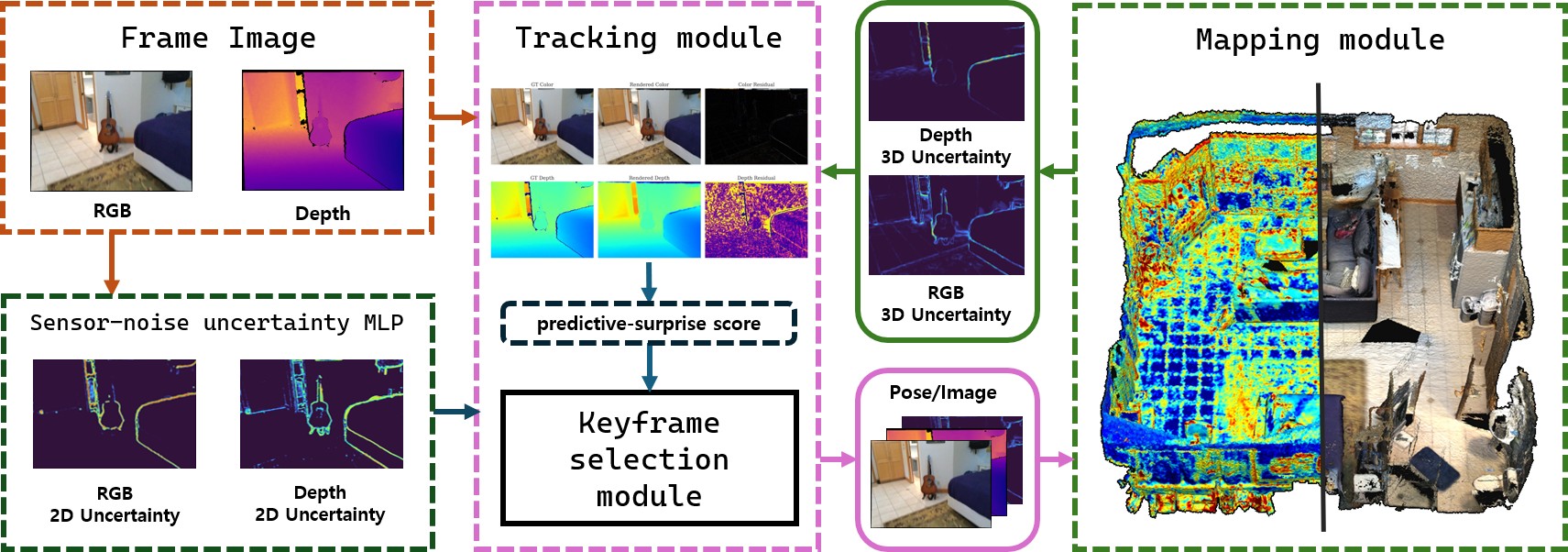}
    \caption{BayesianGS-SLAM framework overview}
    \label{framework}
    \vspace{-15pt}
\end{figure*}



In Sections III-A to III-C, we introduce a tractable predictive-uncertainty formulation and describe its task-specific use in mapping, tracking, and keyframe selection in neural rendering-based SLAM. Section~III-D instantiates the formulation in a 3DGS-based SLAM framework.

\subsection{Predictive Uncertainty Formulation for Mapping}
We introduce a tractable uncertainty formulation motivated by probabilistic reasoning for neural rendering-based SLAM. 
In the mapping stage, we decompose predictive uncertainty into map-representation uncertainty and sensor-noise uncertainty. 
The map-representation uncertainty is modeled as an opacity-induced predictive uncertainty component by propagating opacity variance to the rendered image space, while the sensor-noise uncertainty captures noise and ambiguity in RGB-D measurements. 
Since opacity controls visibility, accumulated transmittance, and rendering weights in Gaussian splatting, opacity-induced uncertainty affects both color and depth predictions.

The goal is to learn a neural field $\mathcal{F}$ given a set of paired images and camera poses, $\mathcal{D} = \{ \{Y_1, \mathcal{P}_1\}, \{Y_2, \mathcal{P}_2\}, \ldots, \{Y_i, \mathcal{P}_i\} \}$, where $Y = \{ y_1, y_2, \ldots \}$ denotes the image observations and $\mathcal{P}$ denotes the camera poses. Given the frame set $\mathcal{D}$, the mapping process can be interpreted as learning a map representation that maximizes the probability of observing $\mathcal{D}$. We define image rendering as a system in which noise is added to an image generated by a 3D rendering model, i.e., $Y = \psi(\mathcal{F}_{\theta}) + \xi$. Intuitively, we assume that the uncertainty inherent in 3D space corresponds to map-representation uncertainty, which becomes larger in less observed regions, whereas the noise introduced when the 3D space is projected onto the 2D image plane corresponds to sensor-noise uncertainty. This can be formulated as follows: 
\begin{align} 
\operatorname*{arg\,max}_{\theta} P_{\theta}(\mathcal{Y}|\mathcal{P}) 
& = \prod_{n,m} P_{\theta}(y_{n,m}|\mathcal{P}_n) \\
\nonumber
& = \prod_{n,m} \int P( y_{n,m} | \psi( \mathcal{F}_\theta)) P(\psi( \mathcal{F}_\theta)) d \mathcal{F}_\theta .
\end{align}
Here, $\theta$ denotes the parameters used to estimate the neural field, $\psi$ is the rendering function, and the neural field $\mathcal{F}_{\theta}$ is obtained as $\mathcal{F}_{\theta}(X_{\mathcal{P}})$ by taking the 3D input space $X$ associated with the camera pose $\mathcal{P}$ as input. The indices $n$ and $m$ denote the frame index and pixel index, respectively. Under our assumption, maximizing the probability of the frame dataset $\mathcal{D}$ can be decomposed into the likelihood term $P( y_{n,m} | \psi( \mathcal{F}_\theta))$, modeled with $\xi \sim \mathcal{N}(0,\sigma_{\mathrm{sensor}}^2)$, and the predictive distribution $P(\psi(F_\theta))$ induced by
map-representation uncertainty.
For notational simplicity, we denote by $P(\psi(\mathcal{F}_{\theta}))$ the predictive distribution in the rendered image space induced by the uncertainty of $\mathcal{F}_{\theta}$. We approximate the uncertainty of the map representation with a Gaussian distribution, which allows predictive uncertainty to be propagated analytically after rendering linearization.

As is well known, the rendering function $\psi$ is nonlinear. Therefore, even if the neural field $\mathcal{F}$ is assumed to follow a Gaussian distribution, the rendered image follows an unknown distribution. Since this prevents the probability distribution from being computed analytically, existing methods typically estimate the parameters using approximate inference techniques, such as variational inference. However, as shown in \cite{lee2025bayesian}, the rendering function can be approximated by a linear function. Following this intuition, we approximate the rendering function through linearization so that the resulting distribution becomes tractable.

Accordingly, we propose a modified linearized rendering approximation when $\mathcal{F}$ follows a Gaussian distribution. Starting from the standard front-to-back alpha-compositing equation $C=\sum_i c_i T_i\alpha_i $, where $T_i=\prod_{j<i}(1-\alpha_j)$. We model uncertainty through opacity because opacity controls visibility and accumulated transmittance in Gaussian splatting, thereby affecting both color and depth predictions. When the opacity $\alpha_i$ follows a Gaussian distribution $\mathcal{N}(\mu_{\alpha_i}, \sigma_{\alpha_i})$, the mean computed by the rendering function of the neural field is given by
\begin{align} 
\mu_{\psi(\mathcal{F}_{\theta})} 
&= \sum_i c_i T_i \mu_{\alpha_i} \\
\nonumber
&\approx \frac{ \sum_i c_i T_i \mu_{\alpha_i} }{ \sum_i  T_i \mu_{\alpha_i} },
\quad \text{where } \sum_i  T_i \alpha_i \approx 1 .
\end{align}
Here, $c_i$ denotes the color of the neural field at the $i$-th sample, and $T_i$ denotes the accumulated transmittance. As discussed in \cite{lee2025bayesian}, when $T_i$ is assumed to be fixed, the rendering equation can be linearized using a first-order Taylor approximation, and the resulting variance is given by
\begin{align} 
\label{randering3}
\operatorname{Var}[\psi(\mathcal{F}_{\theta})] 
&\approx \operatorname{Var}\left[
\sum_i \frac{\partial \psi(\mathcal{F}_{\theta})}{\partial \alpha_i} \alpha_i
\right] \nonumber \\
&= \sum_i 
\left( c_i - \psi(\mathcal{F}_{\theta}) \right)^2 
T_i^2 \sigma_{\alpha_i}^2.
\end{align}
Consequently, the distribution of the rendered image can be approximated as
\begin{align} 
\psi(\mathcal{F}_{\theta}) 
\sim 
\mathcal{N}\left(
\sum_i c_i T_i \mu_{\alpha_i}, 
\sum_i 
\left( c_i - \psi(\mathcal{F}_{\theta}) \right)^2 
T_i^2 \sigma_{\alpha_i}^2
\right).
\end{align}
By assuming that the likelihood term $P_{\theta}( y_{n,m} | \psi( \mathcal{F}_\theta))$ is Gaussian with observation noise $\xi \sim \mathcal{N}(0,\sigma_{\mathrm{sensor}}^2)$, the mapping loss can be written as
\begin{align} 
\label{mapping2}
\theta_{\mathrm{opt}}  
&= \operatorname*{arg\,min}_{\theta} 
\sum_{n,m}
\Bigg[
\frac{
\left( y_{n,m} - \psi( \mathcal{F}_\theta) \right)^2
}{
\sigma_{\mathrm{sensor}}^2 
+ \sum_i 
\left( c_i - \psi(\mathcal{F}_{\theta}) \right)^2 
T_i^2 \sigma_{\alpha_i}^2
} \nonumber \\
&\quad
+ \log \left(
\sigma_{\mathrm{sensor}}^2
+ \sum_i 
\left( c_i - \psi(\mathcal{F}_{\theta}) \right)^2 
T_i^2 \sigma_{\alpha_i}^2
\right)
\Bigg].
\end{align}

As discussed in \cite{kendall2017uncertainties}, the likelihood term $P( y_{n,m} | \psi( \mathcal{F}_\theta))$ can be interpreted as modeling sensor-noise uncertainty, whereas $P(\psi( \mathcal{F}_\theta))$ captures map-representation uncertainty induced by uncertainty in the neural field space. Therefore, to quantify predictive uncertainty, we learn not only the distribution of the neural field, which represents map-representation uncertainty, but also the likelihood distribution, which represents sensor-noise uncertainty. The likelihood distribution, modeled as $\xi \sim \mathcal{N}(0,\sigma_{\mathrm{sensor}}^2)$, is estimated by learning the parameters of a separate MLP.

\subsection{Uncertainty-Normalized Tracking from a Probabilistic Formulation}



In the tracking stage, the camera pose is estimated using the map representation learned during the mapping stage. 
Since tracking requires inferring the camera pose $\mathcal{P}^{\mathrm{new}}$ for a newly observed frame $Y^{\mathrm{new}}$, it can be formulated as a problem of finding the pose $\mathcal{P}^{\mathrm{new}}$ that maximizes the probability of observing $Y^{\mathrm{new}}$:
\begin{align} 
& \operatorname*{arg\,max}_{\mathcal{P}^{\mathrm{new}}}  
P(Y^{\mathrm{new}}|\mathcal{P}^{\mathrm{new}}) \\
\nonumber
& = \prod_m \int 
P(Y^{\mathrm{new}}_m| \psi( \mathcal{F}_\theta^{\mathrm{new}}  )  ) 
P(\psi( \mathcal{F}_\theta^{\mathrm{new}} )) 
d \mathcal{F}_\theta^{\mathrm{new}} .
\end{align}
Here, $\mathcal{F}_\theta^{\mathrm{new}} = \mathcal{F}_\theta(X_{\mathcal{P}^{\mathrm{new}}})$. 
The parameters $\theta$, which estimate the neural field $\mathcal{F}$ and the uncertainty $\sigma_{\theta}$ learned in Section~III-A, are optimized using keyframe data. 
However, due to the sequential nature of SLAM observations, a newly observed frame is expected to belong to a domain similar to that of the learned keyframes. 
Therefore, the neural field evaluated at the new camera pose can be approximated as
$\mathcal{F}_\theta^{\mathrm{new}} = \mathcal{F}_\theta(X_{\mathcal{P}^{\mathrm{new}}}) \approx \mathcal{F}_\theta(X_{\mathcal{P}^{*}})$,
where $\mathcal{P}^{*}$ denotes a nearby keyframe pose.

Consequently, the corresponding negative log-likelihood objective is given by:
\begin{align} 
\label{tracking2}
\mathcal{P}^{\mathrm{new}}_{\mathrm{opt}} 
&= \operatorname*{arg\,min}_{\mathcal{P}^{\mathrm{new}}} 
\sum_m
\Bigg[
\frac{
\left( y^{\mathrm{new}}_{m} - \psi( \mathcal{F}_\theta^{\mathrm{new}}) \right)^2
}{
\sigma_{\theta}^2 
+ \sum_i 
\left( c_i - \psi(\mathcal{F}_\theta^{\mathrm{new}}) \right)^2 
T_i^2 \sigma_{\alpha_i}^2
} \nonumber \\
&\quad
+ \log \left(
\sigma_{\theta}^2
+ \sum_i 
\left( c_i - \psi(\mathcal{F}_\theta^{\mathrm{new}}) \right)^2 
T_i^2 \sigma_{\alpha_i}^2
\right)
\Bigg].
\end{align}


Equation~\eqref{tracking2} provides the probabilistic motivation for using predictive uncertainty during tracking. However, the implemented tracker does not directly optimize this full Gaussian negative log-likelihood. For stable online pose optimization, we instead use a robust uncertainty-normalized L1 surrogate that retains predictive-uncertainty normalization while omitting the log-variance term. The practical tracking objective and its implementation are described in Section~III-D.

\subsection{predictive-surprise-based Keyframe Selection}

Existing neural-rendering-based SLAM systems commonly select keyframes
using fixed temporal intervals or geometric and residual-based
heuristics. Although computationally simple, these strategies may
trigger redundant mapping updates for frames that are already well
explained by the current map. We therefore seek to prioritize
observations that are insufficiently explained by the current
predictive model.

From an information-theoretic perspective, the information gain
associated with a newly observed frame $Y^{\mathrm{new}}$ can be written
as
\begin{align}
\operatorname{IG}(Y^{\mathrm{new}})
&=
D_{\mathrm{KL}}
\left(
P(\mathcal{F}_{\theta}
\mid \mathcal{D}^{*},Y^{\mathrm{new}})
\,\|\, 
P(\mathcal{F}_{\theta}\mid\mathcal{D}^{*})
\right).
\label{eq:exact_ig}
\end{align}
However, evaluating Eq.~\eqref{eq:exact_ig} requires updating the
posterior map distribution for every incoming frame, which is
computationally impractical in online SLAM.

Rather than computing the exact information gain, we use the negative
log posterior-predictive likelihood as a tractable
predictive-surprise score:

\begin{align} 
\nonumber
& \mathcal{S}_{\mathrm{KF}}(Y^{\mathrm{new}}) \\
&\propto - \log P(Y^{\mathrm{new}}| \mathcal{D}^* ) \nonumber \\
&= - \log
\prod_m 
\int 
P(Y^{\mathrm{new}}_m| \psi( \mathcal{F}_\theta^{\mathrm{new}} ) )
P(\psi( \mathcal{F}_\theta^{\mathrm{new}} ))
d \mathcal{F}_\theta^{\mathrm{new}}  \\
\nonumber
&= \sum_m 
- \log 
\int 
P(Y^{\mathrm{new}}_m| \psi( \mathcal{F}_\theta^{\mathrm{new}} ) )
P(\psi( \mathcal{F}_\theta^{\mathrm{new}} ))
d \mathcal{F}_\theta^{\mathrm{new}} .
\end{align}

This quantity measures the predictive surprise of the incoming observation under the current map: a larger value indicates that the frame is less well explained by the current predictive distribution and is therefore more likely to provide a useful mapping update.

A frame is selected as a keyframe when $\mathcal{S}_{\mathrm{KF}}$ exceeds a predefined threshold. In addition, we retain periodic keyframe insertion as a fallback to prevent the map from remaining unchanged for an excessively long interval. Tracking and keyframe selection reuse the same predictive RGB-D uncertainty but employ different task-specific objectives. The tracker uses an uncertainty-normalized L1 objective for robust pose estimation, whereas keyframe selection uses the Gaussian predictive NLL as a predictive-surprise score.

\subsection{BayesianGS-SLAM Framework}

In this section, we present BayesianGS-SLAM, an uncertainty-aware 3DGS SLAM framework based on the probabilistic model of neural rendering-based SLAM derived above. 
As illustrated in Fig.~\ref{framework}, 

our system is built upon the deterministic RGB-D 3DGS SLAM pipeline of LoopSplat~\cite{zhu2025loopsplat}. We retain its 3DGS map representation, rendering-based camera tracking, and online mapping procedure as the base SLAM backbone. On top of this backbone, we introduce predictive uncertainty estimation, uncertainty-normalized tracking, and predictive-surprise-based keyframe selection.
The keyframe selection module determines whether each newly observed frame should be inserted as a keyframe.

In the mapping module, we learn the uncertainty of the 3DGS map representation by constructing a negative log-likelihood objective based on Eq.~\eqref{mapping2}, in addition to the conventional rendering loss. 
Specifically, the predictive uncertainty combines sensor-noise uncertainty predicted from the RGB-D observations and opacity-induced map-representation uncertainty propagated through the renderer.
The map-representation uncertainty is rendered as an uncertainty image using Eq.~\eqref{randering3}, whereas the frame-dependent sensor-noise uncertainty is learned by a separate MLP that takes RGB-D observations as input. 
This follows the intuition of \cite{sandstrom2023uncle}, which aims to capture instantaneous observation noise by learning sensor-noise uncertainty online. 
The uncertainty network is optimized during the mapping stage together with the 3DGS representation, and its parameters are kept fixed during tracking.

For compact notation, we define the predictive variances for color and depth at pixel $(n,m)$ as
\begin{align}
\label{notation_D_mapping}
\Sigma_{C,n,m}
&= \sigma_{\mathrm{sensor},C,n,m}^{2}
+ \sum_i 
\left\lVert c_i - C_{\psi,n,m} \right\rVert_2^2
T_i^2 \sigma_{\alpha_i}^{2}, \\
\label{notation_C_mapping}
\Sigma_{D,n,m}
&= \sigma_{\mathrm{sensor},D,n,m}^{2}
+ \sum_i 
\left( d_i - D_{\psi,n,m} \right)^2
T_i^2 \sigma_{\alpha_i}^{2}.
\end{align}
Here, $\sigma_{\mathrm{sensor},C,n,m}$ and $\sigma_{\mathrm{sensor},D,n,m}$ denote the standard deviations predicted by the uncertainty network for color and depth, respectively.

The final mapping loss is then defined as
\begin{align}
\mathcal{L}_{\mathrm{map}}
&= \mathcal{L}_{\mathrm{3DGS}}
+ \gamma 
\left(
\mathcal{L}_{C}^{\mathrm{map}} 
+ \mathcal{L}_{D}^{\mathrm{map}}
\right), \\
\mathcal{L}_{C}^{\mathrm{map}}
&= \sum_{n,m}
\left[
\frac{
\left\lVert C_{n,m} - C_{\psi,n,m} \right\rVert_2^2
}{
\Sigma_{C,n,m}
}
+ \log \Sigma_{C,n,m}
\right], \\
\mathcal{L}_{D}^{\mathrm{map}}
&= \sum_{n,m}
\left[
\frac{
\left( D_{n,m} - D_{\psi,n,m} \right)^2
}{
\Sigma_{D,n,m}
}
+ \log \Sigma_{D,n,m}
\right],
\end{align}
where we use the conventional 3DGS rendering loss following \cite{yugay2023gaussian}.
Here, $c_i$ and $d_i$ denote the color and depth of the $i$-th 3D Gaussian, respectively. 
In addition, $C$, $D$, $C_\psi$, and $D_\psi$ denote the ground-truth color, ground-truth depth, rendered color, and rendered depth, respectively.

During tracking, the 3DGS and uncertainty-network parameters are fixed.
We reuse the predictive variances in Eq.~\eqref{notation_D_mapping}-\eqref{notation_C_mapping} to normalize the RGB-D
residuals. As discussed in Section III-B, we use a robust L1 surrogate and
omit the log-variance term for stable online pose optimization:
\begin{align} 
\mathcal{L}_{\mathrm{track}} 
&= \lambda_c \mathcal{L}_{C}^{\mathrm{track}} 
+ \lambda_d \mathcal{L}_{D}^{\mathrm{track}}, \\
\mathcal{L}_{C}^{\mathrm{track}} 
&= \sum_{n,m}  
\frac{
\left\lVert C_{n,m} - C_{\psi,n,m} \right\rVert_1
}{
\sqrt{\Sigma_{C,n,m}}
}, \\
\mathcal{L}_{D}^{\mathrm{track}} 
&= \sum_{n,m} 
\frac{
\left| D_{n,m} - D_{\psi,n,m} \right|
}{
\sqrt{\Sigma_{D,n,m}}
}.
\end{align}

In the keyframe selection module, we use both static and dynamic keyframe selection strategies. 
In static keyframe selection, a frame is inserted as a keyframe every $N$ frames after the most recent keyframe. 

In dynamic keyframe selection, we evaluate the Gaussian predictive NLL defined in section~III-C as a predictive-surprise score and insert a new keyframe when the score exceeds a threshold $K$.
The keyframe score reuses the predictive RGB-D uncertainty already available from the framework, avoiding a separate uncertainty-estimation module for keyframe selection.
As a result, our method maintains SLAM performance while using fewer keyframes than conventional approaches.

\section{Experiment}


\subsection{Implementation and Setup}



We evaluate BayesianGS-SLAM on the TUM~\cite{sturm2012benchmark} and ScanNet~\cite{dai2017scannet} RGB-D datasets using an NVIDIA RTX 4090. We set $\gamma=10^{-3}$, $\lambda_c=\lambda_d=900$, $N=5$, and $K=0$. The sensor-noise uncertainty network takes RGB-D observations as input and consists of four convolutional and four fully connected layers. The minimum sensor-noise and map-representation uncertainty values are set to $10^{-4}$ and $10^{-3}$, respectively. We evaluate tracking using ATE RMSE, rendering using PSNR, SSIM, and LPIPS, uncertainty--error ranking using AUSE MAE, and efficiency using the number of selected keyframes, mapping calls, and FPS.

\begin{table}[t]
\centering
\caption{Tracking Evaluation on TUM \cite{sturm2012benchmark}}
\label{tab:trajectory_results}
\resizebox{\columnwidth}{!}{
\begin{tabular}{lcccccc}
\toprule
\textbf{Method}  & fr1/desk & fr1/desk2 & fr1/room & fr2/xyz & fr3/off. & \textbf{Avg.} \\
\midrule
\rowcolor{gray!15} \multicolumn{7}{l}{\textbf{Neural Implicit Fields}} \\
NICE-SLAM \cite{zhu2022nice}  & 4.26 & 4.99 & 34.49 & 6.19 & 3.87 & 10.76 \\
MIPS-Fusion \cite{tang2023mips}  & 3.00 & N/A & N/A & 1.4 & 4.6 & N/A \\
Point-SLAM \cite{sandstrom2023point}  & 4.34 & 4.54 & 30.92 & 1.31 & 4.8 & 8.92 \\
ESLAM \cite{johari2023eslam}  & 2.47 & 3.69 & 29.73 & \third{1.11} & 2.42 & 7.89 \\
Co-SLAM \cite{wang2023co}  & 2.40 & N/A & N/A & 1.70 & 2.40 & N/A \\
GO-SLAM \cite{zhang2023go}  & \best{1.50} & N/A & \best{4.64} & \second{0.60} & \second{1.30} & N/A \\
Loopy-SLAM \cite{liso2024loopy}  & 3.79 & \second{3.38} & 7.03 & 1.62 & 3.41 & 3.85 \\
\midrule
\rowcolor{gray!15} \multicolumn{7}{l}{\textbf{3D Gaussian Splatting}} \\
SplaTAM \cite{keetha2024splatam}  & 3.35 & 6.54 & 11.13 & 1.24 & 5.16 & 5.48 \\
MonoGS \cite{matsuki2024gaussian}  & \second{1.59} & 7.03 & 8.55 & 1.44 & \third{1.49} & 4.02 \\
Gaussian-SLAM \cite{yugay2023gaussian}  & 2.73 & 6.03 & 14.92 & 1.39 & 5.31 & 6.08 \\
*Photo-SLAM \cite{huang2024photo}  & 2.60 & N/A & N/A & \best{0.35} & \best{1.00} & N/A \\
LoopSplat \cite{zhu2025loopsplat}  & 2.08 & 3.54 & {6.24} & 1.58 & 3.22 & \third{3.33} \\
CG-SLAM \cite{hu2024cg}  & 2.43 & 4.54 & 9.39 & 1.20 & 2.45 & {4.0} \\
VarSplat \cite{tran2026varsplat}   & \third{1.80} & \third{3.40} & \second{6.05} & 1.41 & 3.36 & \second{3.20} \\
\textbf{BayesianGS}  & {2.01} & \best{3.03} & \third{6.23} & 1.43 & 2.91 & \best{3.12} \\
\bottomrule
\end{tabular}
}
\vspace{-10pt}
\end{table}

\begin{table}[t]
\centering
\caption{Tracking Evaluation on ScanNet \cite{dai2017scannet} }
\label{tab:sequence_results}
\resizebox{\columnwidth}{!}{
\begin{tabular}{l c c c c c c c}
\toprule
\textbf{Method} & 00 & 59 & 106 & 169 & 181 & 207 & \textbf{Avg.} \\
\midrule
\rowcolor{gray!15} \multicolumn{8}{l}{\textbf{Neural Implicit Fields}} \\
Co-SLAM \cite{wang2023co}  & 7.1 & 11.1 & 9.4 & \best{5.9} & 11.8 & 7.1 & 8.7 \\
NICE-SLAM \cite{zhu2022nice}  & 12.0 & 14.0 & 7.9 & 10.9 & 13.4 & \third{6.2} & 10.7 \\
ESLAM \cite{johari2023eslam}  & 7.3 & 8.5 & 7.5 & \second{6.5} & 9.0 & \second{5.7} & 7.4 \\
Point-SLAM \cite{sandstrom2023point}  & 10.2 & 10.8 & 8.7 & 7.2 & 22.2 & 14.8 & 12.3 \\
GO-SLAM \cite{zhang2023go}  & \third{5.4} & 7.5 & \second{7.0} & 7.0 & \best{6.8} & 6.9 & \third{6.8} \\
Loopy-SLAM \cite{liso2024loopy}  & \best{4.2} & 7.5 & 8.3 & 7.5 & 10.6 & 7.9 & 7.7 \\
Uni-SLAM \cite{wang2025uni} & 6.1 & 7.8 & 7.4 & \third{5.8} & 9.8 & 5.2 & 7.0 \\
\midrule
\rowcolor{gray!15} \multicolumn{8}{l}{\textbf{3D Gaussian Splatting}} \\
MonoGS \cite{matsuki2024gaussian}  & 9.8 & 32.1 & 8.9 & 10.7 & 21.8 & 7.9 & 15.2 \\
SplaTAM \cite{keetha2024splatam}  & 10.1 & 17.7 & 11.7 & 7.5 & 5.6 & 7.5 & 10.0 \\
Gaussian-SLAM \cite{yugay2023gaussian}  & 21.2 & 12.8 & 13.5 & 13.6 & 21.0 & 13.4 & 15.9 \\
LoopSplat \cite{zhu2025loopsplat}   & 6.2 & \third{7.1} & 7.4 & 10.6 & \second{8.5} & 6.6 & 7.7 \\
CG-SLAM \cite{hu2024cg}  & 7.1 & 7.5 & 8.9 & 8.2 & 11.6 & \best{5.3} & 8.1 \\
VarSplat \cite{tran2026varsplat}  & \second{4.9} & \second{5.8} & \best{6.7} & 6.7 & 8.9 & 6.2 & \best{6.5} \\
\textbf{BayesianGS }   & 5.5 & \best{5.2} & \third{7.1} & 7.8 & \third{8.0} & 6.7 & \second{6.7} \\
\bottomrule
\end{tabular}
}
\vspace{-5pt}
\end{table}

\begin{table}[h] 
\centering
\caption{Rendering Evaluation on 2 datasets.}
\label{tab:rendering_3datasets}
\resizebox{\columnwidth}{!}{
\begin{tabular}{l ccc ccc}
\toprule
\textbf{Dataset} & \multicolumn{3}{c}{\textbf{TUM} \cite{sturm2012benchmark}} & \multicolumn{3}{c}{\textbf{ScanNet} \cite{dai2017scannet} } \\
\cmidrule(lr){2-4} \cmidrule(lr){5-7} 
\textbf{Method} & PSNR $\uparrow$ & SSIM $\uparrow$ & LPIPS $\downarrow$ & PSNR $\uparrow$ & SSIM $\uparrow$ & LPIPS $\downarrow$ \\
\midrule
NICE-SLAM \cite{zhu2022nice} & 14.86 & 0.614 & 0.441 & 17.54 & 0.621 & 0.548 \\
ESLAM \cite{johari2023eslam} & 15.26 & 0.478 & 0.569 & 15.29 & 0.658 & 0.488 \\
Point-SLAM \cite{sandstrom2023point}  & 16.62 & 0.696 & 0.526 & \second{19.82} & \third{0.751} & 0.514 \\
Loopy-SLAM \cite{liso2024loopy}  & 12.94 & 0.489 & 0.645 & 15.23 & 0.629 & 0.671 \\
SplaTAM \cite{keetha2024splatam} & \second{22.80} & \best{0.893} & \best{0.178} & \third{19.14} & 0.716 & \best{0.358} \\
\color{graytext} Gaussian-SLAM \cite{yugay2023gaussian}  & \color{graytext} 25.05 & \color{graytext} 0.929 & \color{graytext} 0.168 & \color{graytext} 27.67 & \color{graytext} 0.923 & \color{graytext} 0.248 \\
LoopSplat \cite{zhu2025loopsplat}  & 22.72 & {0.873} & {0.259} & \best{24.92} & \second{0.845} & \third{0.425} \\
VarSplat \cite{tran2026varsplat} & \third{23.14}   & \third{0.883} & \third{0.248} & \best{24.92} & \best{0.848} & \second{0.422} \\
\textbf{BayesianGS } & \best{23.30}   & \second{0.891} & \second{0.239} & \second{24.88} & \second{0.845} & 0.427 \\
\midrule
\multicolumn{7}{l}{ \quad \quad \  \textcolor{graytext}{Gray indicates evaluation on submaps rather than global map.}} \\
\bottomrule
\end{tabular}
}
\end{table}



\begin{table*}[h] 
\centering
\caption{\textbf{Color} / \textbf{Depth} Uncertainty Evaluation (AUSE MAE $\downarrow$) on 2 datasets}
\label{tab:AUSE_MAE}
\resizebox{\textwidth}{!}{
\begin{tabular}{l l  ccc cccccc c }
\toprule
\textbf{Dataset} & & \multicolumn{3}{c}{\textbf{TUM}  }  & \multicolumn{6}{c}{\textbf{ScanNet} } & \textbf{\multirow{2}{*}{\textbf{Avg}}} \\
\cmidrule(lr){3-5} \cmidrule(lr){6-11} 
\textbf{Method} & & fr1/desk & fr2/xyz & fr3/office  & 0000 & 0059 & 0106 & 0169 & 0181 & 0207 &  \\
\midrule
CG-SLAM \cite{hu2024cg}  & \multirow{4}{*}{\textbf{Color AUSE MAE $\downarrow$}} & {0.0770} & {0.0687} & {0.0724} & {0.0373} & {0.0627} & {0.1027} & {0.0486} & {0.0640} & {0.0367} & {0.0633} \\
VarSplat \cite{tran2026varsplat}  &   & \textbf{0.0117} & {0.0132} & \textbf{0.0110} & {0.0127} & \textbf{0.0153} & \textbf{0.0129} & {0.0081} & {0.0147} & {0.0107} & {0.0123} \\
FisherRF* \cite{jiang2023fisherrf} &   & {0.0280} & {0.0183} & {0.0230} & {0.0213} & {0.0316} & {0.0293} & {0.0128} & {0.0260} & {0.0175} & {0.0231} \\
\textbf{BayesianGS} &                                                        & {0.0118} & \textbf{0.0108} & {0.0127} & \textbf{0.0116} & {0.0155} & {0.0167} & \textbf{0.0073} & \textbf{0.0124} & 
\textbf{0.0101} & \textbf{0.0121} \\
\midrule
CG-SLAM \cite{hu2024cg} & \multirow{4}{*}{\textbf{Depth AUSE MAE $\downarrow$}} & {0.1948} & {0.3355} & {0.1535} & {0.0610} & {0.1359} & {0.3590} & {0.2089} & {0.1338} & {0.0752} & {0.1842} \\ 
VarSplat \cite{tran2026varsplat}   &   & {0.2145} & {1.0522} & {0.1398} & {0.0922} & {0.2140} & {0.2791} & {0.2386} & {0.1116} & {0.1352} & {0.2752} \\
FisherRF* \cite{jiang2023fisherrf} &   & \textbf{0.0567} & {0.5822} & {0.1087} & {0.0599} & {0.2217} & {0.3261} & {0.2405} & {0.1296} & {0.0766} & {0.2002} \\
\textbf{BayesianGS} &                                                        & {0.0580} & \textbf{0.2541} & \textbf{0.0478} & \textbf{0.0331} & \textbf{0.1048} & \textbf{0.2059} & \textbf{0.1401} & \textbf{0.0477} & \textbf{0.0628} & \textbf{0.1060} \\
\midrule
\multicolumn{12}{l}{
\footnotesize{$^{*}$FisherRF is not a SLAM system; we apply its uncertainty formulation to 3DGS maps reconstructed using the estimated poses from our SLAM pipeline.}} \\
\bottomrule
\end{tabular}
}
\vspace{-5pt}
\end{table*}

\begin{table*}[t]
\centering
\caption{Keyframe and mapping efficiency on tum \cite{sturm2012benchmark} }
\label{tab:system_efficiency}
\begin{tabular}{l cc cc cc}
\toprule
\textbf{Dataset} & \multicolumn{2}{c}{\textbf{fr1/desk} } & \multicolumn{2}{c}{\textbf{fr2/xyz} } & \multicolumn{2}{c}{\textbf{fr3/office} } \\
\cmidrule(lr){2-3} \cmidrule(lr){4-5} \cmidrule(lr){6-7} 
\textbf{Method} & \textbf{KF Insertion} & \textbf{Mapping Freq}& \textbf{KF Insertion} & \textbf{Mapping Freq}& \textbf{KF Insertion} & \textbf{Mapping Freq}  \\
\midrule
Gaussian-SLAM \cite{yugay2023gaussian} & Dense ($591$) & Per-frame ($591$)  & Dense ($3393$) & Per-frame ($3393$)  & Dense ($2514$) & Per-frame ($2514$)  \\
LoopSplat \cite{zhu2025loopsplat}     & Dense ($591$) & Per-frame ($591$)  & Dense ($3393$) & Per-frame ($3393$)  & Dense ($2514$) & Per-frame ($2514$)  \\
VarSplat  \cite{tran2026varsplat}    & Dense ($591$) & Per-frame ($591$)   & Dense ($3393$) & Per-frame ($3393$)  & Dense ($2514$) & Per-frame ($2514$) \\
\midrule
\textbf{BayesianGS}      & \textbf{{Sparse} ($137$)} & \textbf{{Keyframe-only} ($137$)}  & \textbf{{Sparse} ($744$)} & \textbf{{Keyframe-only} ($744$)}  & \textbf{{Sparse} ($545$)} & \textbf{{Keyframe-only} ($545$)} \\
\bottomrule
\vspace{-10pt}
\end{tabular}

\end{table*}

\begin{table*}[h]
\centering
\caption{Tracking ATE for different pose-optimization objectives on TUM.}
\label{tab:rebutal_tracking}
\begin{tabular}{lcccccc}
\toprule
\textbf{Method} &  fr1/desk & fr1/desk2 & fr1/room & fr2/xyz & fr3/off. & \textbf{Avg.} \\
\midrule
\textbf{Conventional L1}& {2.17} $\pm$ 0.01 & {4.64} $\pm$ 0.38 & X & 1.43 $\pm$ 0.02  & \textbf{2.98} $\pm$ \textbf{0.23} & X \\
\textbf{Gaussian NLL}& 21.23 $\pm$ 3.21 & 49.50 $\pm$ 8.42 & 68.65 $\pm$ 10.52 & 1.75 $\pm$ 0.08 & 71.05 $\pm$ 16.57 & 42.44 \\
\textbf{Ours uncertainty-normalized loss with log}& 20.40 $\pm$ 1.20 & 56.93 $\pm$ 1.37 & 98.19 $\pm$ 18.58 & 1.71 $\pm$ 0.10 & 71.18 $\pm$ 4.41 & 49.68 \\
\textbf{Ours uncertainty-normalized loss}& \textbf{1.92$\pm$0.10}  & \textbf{3.25$\pm$0.32 }& \textbf{11.32$\pm$9.42 } & \textbf{1.41$\pm$0.01}   & {3.01$\pm$0.14} & {4.18}\\
\bottomrule
\end{tabular}
\vspace{-15pt}
\end{table*}

\begin{figure*}[hbt!]
    \centering
    \begin{subfigure}[c]{0.49\textwidth}
        \centering
        \includegraphics[width=\textwidth]{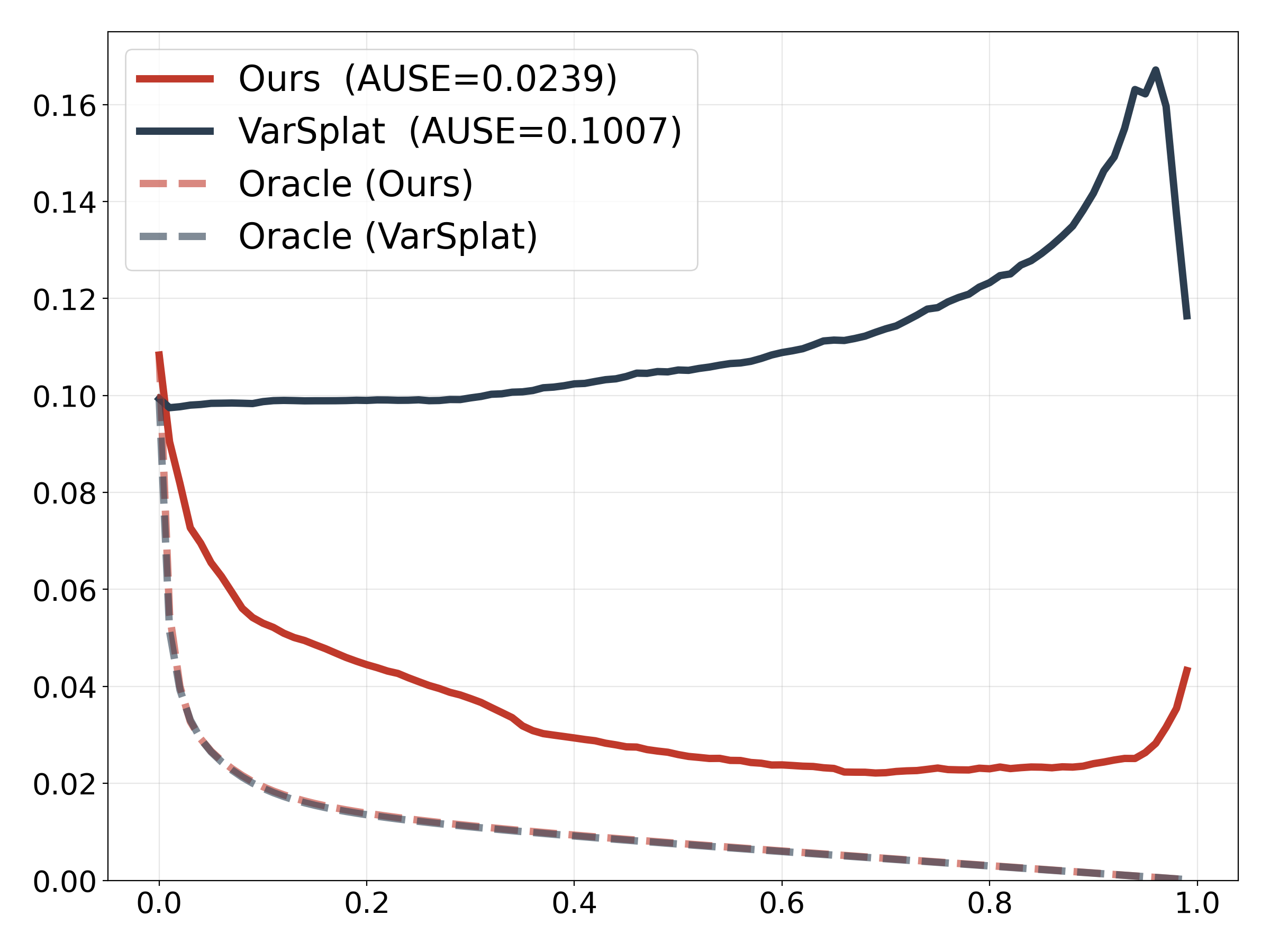}
        \vspace{-20pt}
        \caption{\small AUSE MAE Plot }
        \label{fig:remapping_gt}
    \end{subfigure}
    \begin{minipage}[c]{0.46\textwidth}
        \begin{subfigure}[b]{0.49\linewidth}
            \centering
            \includegraphics[width=\textwidth]{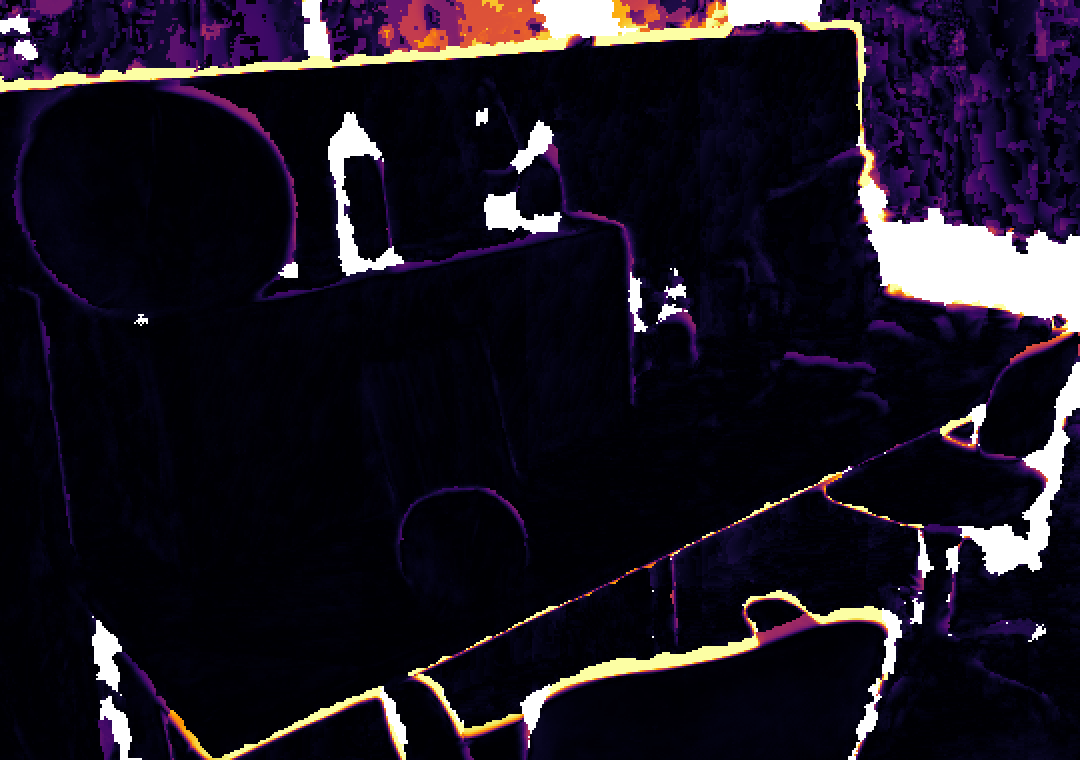}
            \vspace{-15pt}
            \caption{\small VarSplat Residual  }
            \label{fig:remapping_plot}
        \end{subfigure}
        \begin{subfigure}[b]{0.49\linewidth}
            \centering
            \includegraphics[width=\textwidth]{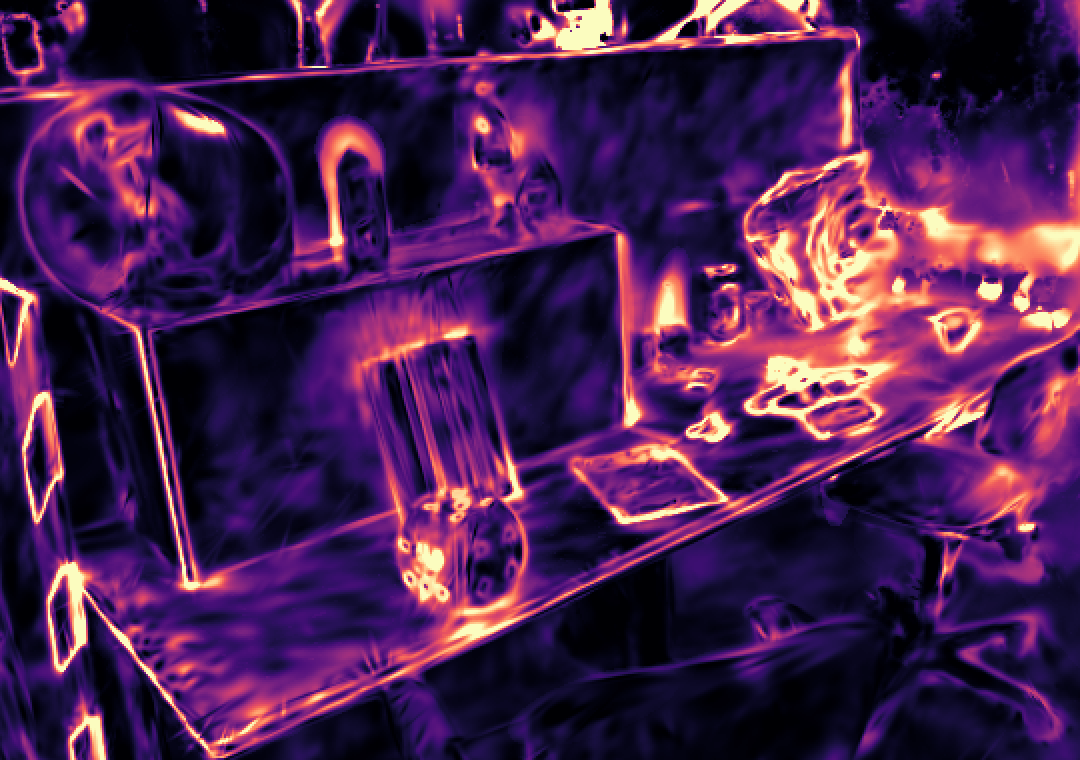}
            \vspace{-15pt}
            \caption{\small VarSplat Uncertainty}
            \label{fig:remapping_est}
        \end{subfigure}

        \begin{subfigure}[b]{0.49\linewidth}
            \centering
            \includegraphics[width=\textwidth]{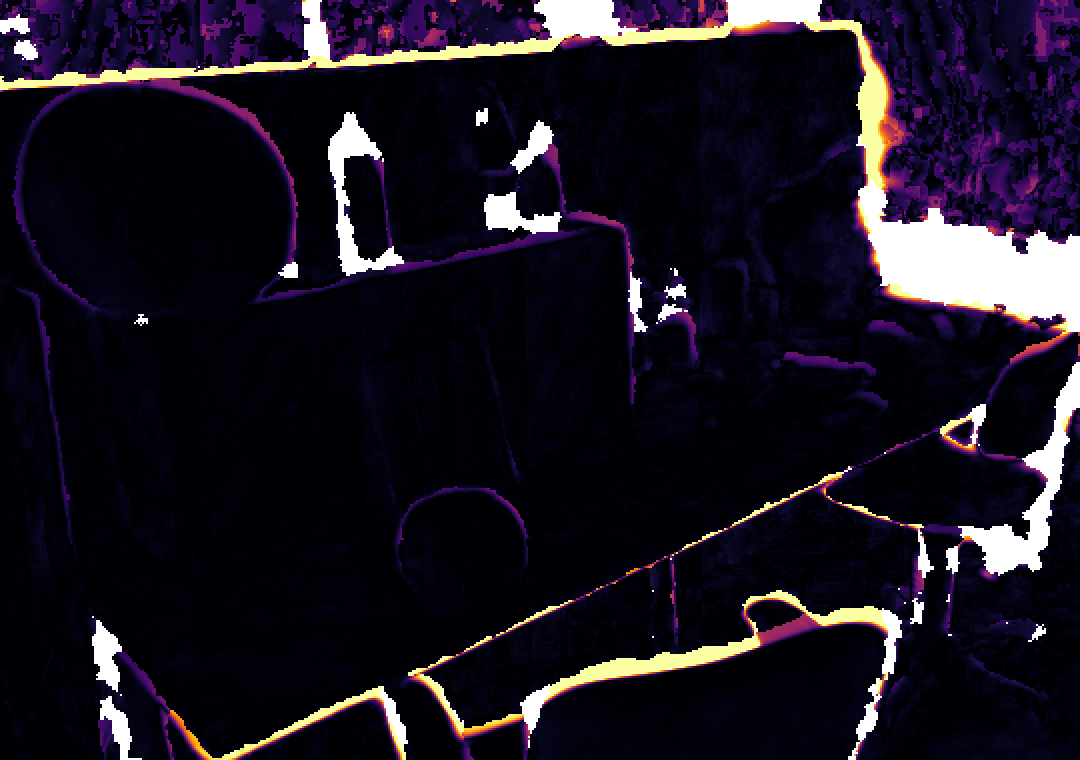}
            \caption{\small BayesianGS Residual  }
            \label{fig:remapping_err}
            \vspace{-15pt}
        \end{subfigure}
        \begin{subfigure}[b]{0.49\linewidth}
            \centering
            \includegraphics[width=\textwidth]{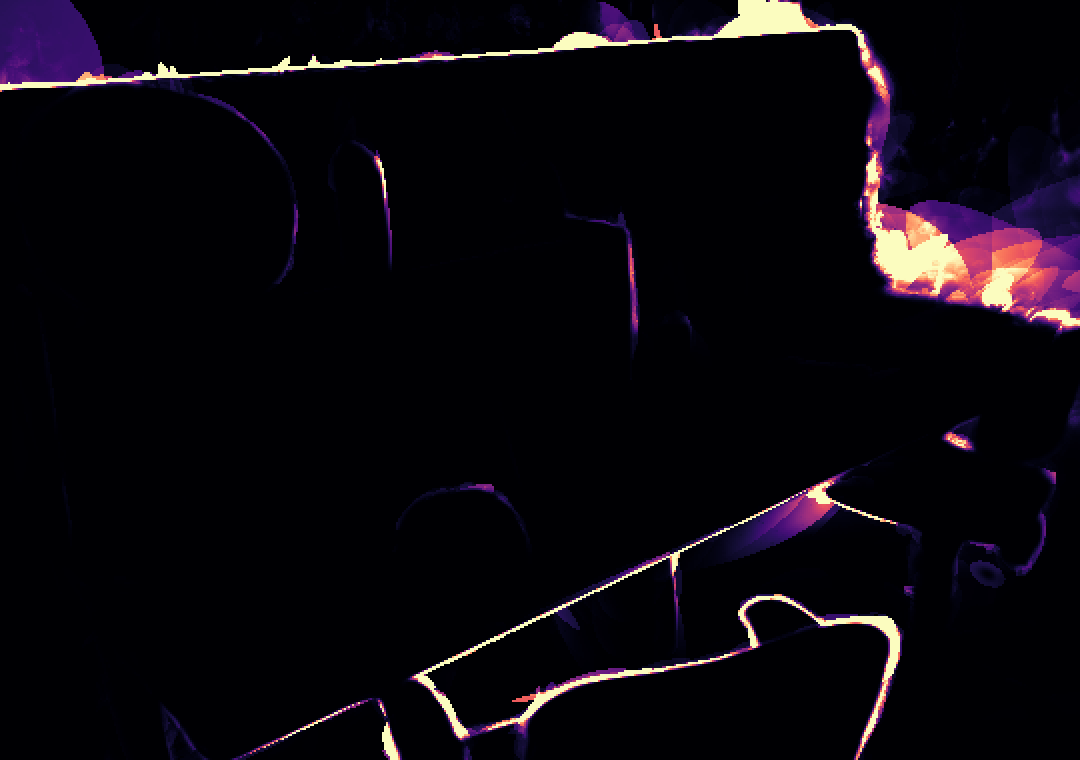}
            \caption{\small BayesianGS Uncertainty}
            \label{fig:remapping_uncert}
            \vspace{-15pt}
        \end{subfigure}
    \end{minipage}
    
    \caption{\small Depth uncertainty evaluation on frame 2324 of TUM fr3/office. 
Residual maps show absolute depth error, and uncertainty maps show rendered depth uncertainty.}
    \label{fig:uncert_result}
    \vspace{-10pt}
\end{figure*}

\begin{figure}[hbt!]
    \centering
    \includegraphics[width=8.0cm, height=4.0cm]{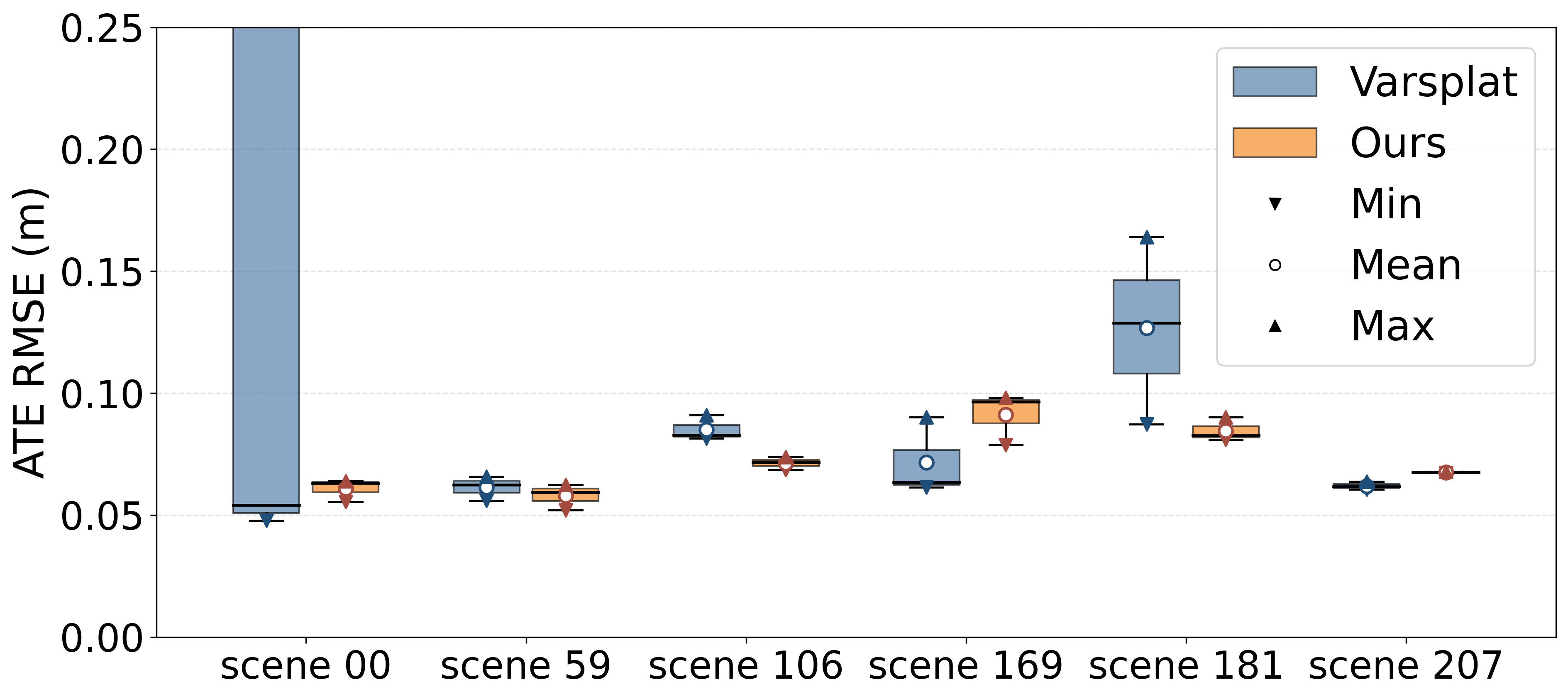}
    \caption{Tracking Variability on ScanNet \cite{dai2017scannet} }
    \label{fig:robustness}
    \vspace{-10pt}
\end{figure}


\begin{figure}[hbt!]
    \centering
    \includegraphics[width=0.45\textwidth]
    {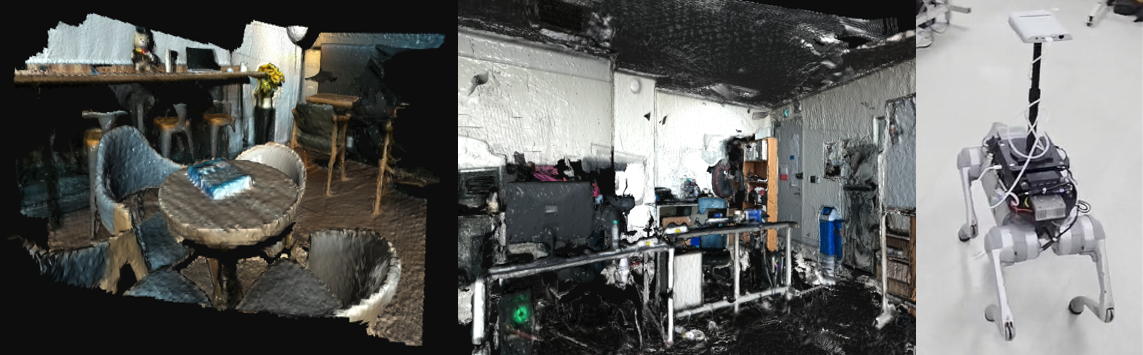}
    \caption{Real-robot demonstration. Left to right: handheld result, robot-mounted result, and experimental robot platform.}
    \label{fig:realworld}
    \vspace{-15pt}
\end{figure}


\begin{table}[t]
\centering
\caption{Module ablation results on TUM \cite{sturm2012benchmark}}
\label{tab:uncertainty_ablation}
\begin{tabular}{ l c c}
\toprule
\textbf{Metric} & \textbf{w/o Track.+ KFS} & \textbf{w/ Track.+KFS} \\
\midrule
ATE RMSE $\downarrow$   & 3.47  &  \textbf{3.12} \\
PSNR $\uparrow$  &  22.64 & \textbf{22.87} \\
Color/Depth AUSE  $\downarrow$  &  0.0136 / 0.1426 & \textbf{0.0132 }/ \textbf{0.1253}  \\
Number of KF $\downarrow$  & 1689.8  & \textbf{380.6} \\
\bottomrule
\end{tabular}
\vspace{-5pt}
\end{table}

\begin{table}[t]
\centering
\caption{FPS $\uparrow$ evaluation on tum \cite{sturm2012benchmark}}
\label{tab:runtime}
\resizebox{\columnwidth}{!}{
\begin{tabular}{l c c c c c}
\toprule
\textbf{Sequence} & UC & fr1/desk & fr2/xyz & fr3/office & Avg.   \\
\midrule
Point-SLAM  \cite{sandstrom2023point}            &  \xmark  & 0.26        & 0.27    & 0.23   & 0.25 \\
Loopy-SLAM  \cite{liso2024loopy}            &  \xmark  & 0.24        & 0.23    & 0.20   & 0.22           \\
SplaTAM   \cite{keetha2024splatam}              &  \xmark  & 0.39        & \third{0.42}     & 0.40  & 0.41           \\
LoopSplat \cite{zhu2025loopsplat}              &  \xmark  & \best{0.59}        & \best{0.59}    & \second{0.59}   & \best{0.59}        \\
VarSplat  \cite{tran2026varsplat}             &  \cmark  & \third{0.44}        & 0.41    & \third{0.52}   & \third{0.47}               \\
\textbf{BayesianGS}       &  \cmark  & \second{0.50}        & \second{0.55}    & \best{0.64}   & \second{0.56}      \\
\bottomrule
\end{tabular}
}
\vspace{-15pt}
\end{table}

\subsection{Evaluation on Tracking}

Tables~\ref{tab:trajectory_results} and~\ref{tab:sequence_results} report the tracking results. BayesianGS-SLAM achieves the lowest average ATE on TUM (3.12 cm) and the second-lowest on ScanNet (6.7 cm). Fig.~\ref{fig:robustness} further evaluates run-to-run variation over three runs on ScanNet, where our method shows lower variation than VarSplat.

\subsection{Evaluation on Rendering and Uncertainty}

Table~\ref{tab:rendering_3datasets} reports rendering performance. Following the protocol of prior GS-based SLAM work~\cite{zhu2025loopsplat}, rendering quality is evaluated on the keyframes used for training. Therefore, these results measure system-level training-view reconstruction fidelity rather than a controlled cross-method novel-view comparison. BayesianGS-SLAM maintains competitive rendering quality on both datasets.

Table~\ref{tab:AUSE_MAE} evaluates uncertainty error ranking using AUSE MAE. In addition to uncertainty-aware SLAM baselines, we include FisherRF, whose uncertainty formulation is applied to 3DGS maps reconstructed using poses estimated by our SLAM pipeline. BayesianGS-SLAM achieves an average Color AUSE of 0.0121, comparable to VarSplat, and the lowest average Depth AUSE of 0.1060, compared with 0.1842 for CG-SLAM, 0.2752 for VarSplat, and 0.2002 for FisherRF. These results indicate improved uncertainty-error ranking, particularly for depth, rather than absolute uncertainty calibration. We attribute the stronger depth uncertainty--error ranking to the opacity-induced map-representation component, which directly reflects changes in visibility and compositing weights shared by depth rendering. Fig.~\ref{fig:uncert_result} provides a qualitative comparison of depth residuals and uncertainty maps.

\subsection{Evaluation on Keyframe Selection}

In this section, we evaluate the efficiency of the proposed keyframe selection module, as shown in Table~\ref{tab:system_efficiency}. 
We measure the number of inserted keyframes and the number of mapping module calls on three TUM sequences: fr1/desk, fr2/xyz, and fr3/office. 
Our method reduces both the number of keyframes and the number of mapping module calls by approximately 70\%, while maintaining competitive tracking and mapping performance.

This efficiency also provides a practical benefit in terms of runtime. 
As shown in Table~\ref{tab:runtime}, the FPS measured on the TUM dataset demonstrates that our method runs faster than VarSplat and most compared methods.
Although LoopSplat achieves the highest FPS, it does not include an uncertainty estimation module. 
Our method maintains competitive tracking and rendering performance while substantially reducing redundant mapping updates, despite the additional per-frame cost of uncertainty estimation. However, reducing the mapping frequency may delay map refinement when the keyframe criterion fails to identify an informative observation.


\subsection{Ablation Studies}

In the ablation studies, we analyze the effect of the proposed uncertainty-aware tracking and keyframe selection modules on tracking performance. 
We further show that the proposed uncertainty-normalized L1 surrogate achieves substantially lower tracking error than the Gaussian NLL and log-term variants while successfully tracking all evaluated sequences.
In addition, to demonstrate the applicability of our framework to real robotic systems, we present mapping results obtained from data collected using handheld and robot-mounted sensors.

\subsubsection{Effect of Tracking and Keyframe Selection Modules}

To evaluate the effect of the proposed uncertainty-aware tracking and keyframe selection modules, we compare the baseline configuration without these modules against the full configuration with both modules enabled. 
We report ATE RMSE, PSNR, color/depth AUSE MAE, and the number of selected keyframes averaged over five TUM scenes, as shown in Table~\ref{tab:uncertainty_ablation}. 
The full configuration improves tracking accuracy, rendering quality, and uncertainty estimation while substantially reducing the number of keyframes. 
These results indicate that the proposed tracking and predictive-surprise-based keyframe selection are effective.

\subsubsection{Effect of Tracking loss}
We further evaluate the tracking surrogate in Table~\ref{tab:rebutal_tracking}. Results are averaged over three independent runs. The proposed uncertainty-normalized L1 objective successfully tracks all five sequences and achieves an average ATE of 4.18 cm, whereas conventional L1 fails on \texttt{fr1/room}. Direct optimization of the Gaussian NLL and the uncertainty-normalized objective with the log term results in substantially larger average ATEs of 42.44 cm and 49.68 cm, respectively. These results support the use of uncertainty-normalized L1 as a practical robust surrogate for online pose optimization rather than an exact likelihood objective.

\subsubsection{Evaluation with Real-Robot Data}



We further provide qualitative results on robot-collected RGB-D sequences, including handheld and robot-mounted settings, we present qualitative mapping results in Fig.~\ref{fig:realworld}.
The results show that the proposed framework can be applied to robot-collected RGB-D data and can reconstruct consistent map representations in practical indoor environments. Additional results are provided in the supplementary video.

\section{Conclusion}


In this paper, we presented BayesianGS-SLAM, an uncertainty-aware 3DGS SLAM framework that estimates predictive uncertainty for both color and depth. The proposed formulation combines sensor-noise uncertainty with opacity-induced map-representation uncertainty and reuses the resulting predictive uncertainty for mapping, uncertainty-normalized tracking, and predictive-surprise-based keyframe selection. Experiments on TUM and ScanNet demonstrate improved depth uncertainty--error ranking and competitive tracking and rendering performance while substantially reducing redundant mapping updates. Qualitative experiments on robot-collected RGB-D data further demonstrate the applicability of the proposed framework beyond benchmark datasets.

\bibliographystyle{IEEEtran}
\bibliography{bib}

\end{document}